\pdfoutput=1

\documentclass[11pt]{article}

\usepackage{acl}

\usepackage{times}
\usepackage{latexsym}

\usepackage[T1]{fontenc}

\usepackage[utf8]{inputenc}

\usepackage{microtype}

\usepackage[textsize=tiny,textwidth=2cm]{todonotes}
\makeatletter
\newcommand*\iftodonotes{\if@todonotes@disabled\expandafter\@secondoftwo\else\expandafter\@firstoftwo\fi}  
\makeatother

\usepackage{amsmath}
\usepackage{amssymb}
\usepackage{multirow}
\usepackage{subfig}
\usepackage{xspace}
\newcommand{\tabincell}[2]{\begin{tabular}{@{}#1@{}}#2\end{tabular}}

\newcommand{\dec}{\ensuremath{\mathtt{Decoder}}\xspace}
\newcommand{\transformer}{\ensuremath{\mathtt{Transformer}}\xspace}

\title{Clues Before Answers: Generation-Enhanced Multiple-Choice QA}

\author{Zixian Huang \and Ao Wu \and Jiaying Zhou \\
  State Key Laboratory for Novel Software Technology, Nanjing University, Nanjing, China \\
  \texttt{\{zixianhuang,awu,jyzhou\}@smail.nju.edu.cn} \\
  \AND
  Yu Gu \\
  The Ohio State University, Columbus, USA \\
  \texttt{gu.826@osu.edu} \\
  \AND
  Yue Zhao \and Gong Cheng \\
  State Key Laboratory for Novel Software Technology, Nanjing University, Nanjing, China \\
  \texttt{yuezhao@smail.nju.edu.cn, gcheng@nju.edu.cn}
}

\begin{document}
\maketitle
\begin{abstract}
A trending paradigm for multiple-choice question answering~(MCQA) is using a text-to-text framework. By unifying data in different tasks into a single text-to-text format, it trains a generative encoder-decoder model which is both powerful and universal. However, a side effect of twisting a generation target to fit the classification nature of MCQA is the under-utilization of the decoder and the knowledge that can be decoded. To exploit the generation capability and underlying knowledge of a pre-trained encoder-decoder model, in this paper, we propose a generation-enhanced MCQA model named GenMC. It generates a clue from the question and then leverages the clue to enhance a reader for MCQA. It outperforms text-to-text models on multiple MCQA datasets.
\end{abstract}

\section{Introduction}
\label{sec:introduction}

Multiple-choice question answering~(MCQA)
aims at selecting the correct answer from a set
of options given a question. This long-standing challenge in natural language processing (NLP) requires machines to have a wealth of knowledge, such as commonsense knowledge~\cite{CSQA,OBQA} and scientific knowledge~\cite{ARC,QASC,DBLP:conf/emnlp/HuangSLWCZDQ19,DBLP:conf/aaai/LiS021}, and have reasoning skills such as multi-hop reasoning~\cite{GapQA} and logical reasoning~\cite{DBLP:conf/iclr/YuJDF20,DBLP:conf/ijcai/LiuCLHWZ20,DBLP:journals/corr/abs-2203-08992}.

MCQA has made great progress with the development of pre-trained language models~(PLMs). Basically there are two types of PLMs that are suitable for different tasks. BERT~\cite{BERT} and its variants such as RoBERTa~\cite{RoBERTa} and ALBERT~\cite{ALBert} are encoder-only models, being more suitable for natural language understanding~(NLU) tasks including MCQA and other classification and regression tasks. T5~\cite{T5} and BART~\cite{BART} are encoder-decoder models, being more suitable for natural language generation~(NLG) tasks. However, encoder-decoder models can also be applied to MCQA~\cite{UnifiedQA,CALM}. This is enabled by the text-to-text framework, which transforms data in different tasks into a unified text-to-text format so that knowledge spanning many and various tasks can be learned, aggregated, and used by a single model.

\paragraph{Research Question}
To fit MCQA, existing implementations of the text-to-text framework take all the options as input and are trained to generate one of the options, i.e., to copy some tokens from the input. However, this is inconsistent with how encoder-decoder models are pre-trained so that their underlying knowledge may not be sufficiently exploited. Indeed, \citet{EncT5} have found that in classification and regression tasks, the decoder layer is often under-utilized. One research question is \emph{how to apply pre-trained encoder-decoder models in a more natural way to MCQA}, in particular, to exploit their NLG capabilities.

\begin{figure}
    \centering
    \includegraphics[width=0.9\columnwidth]{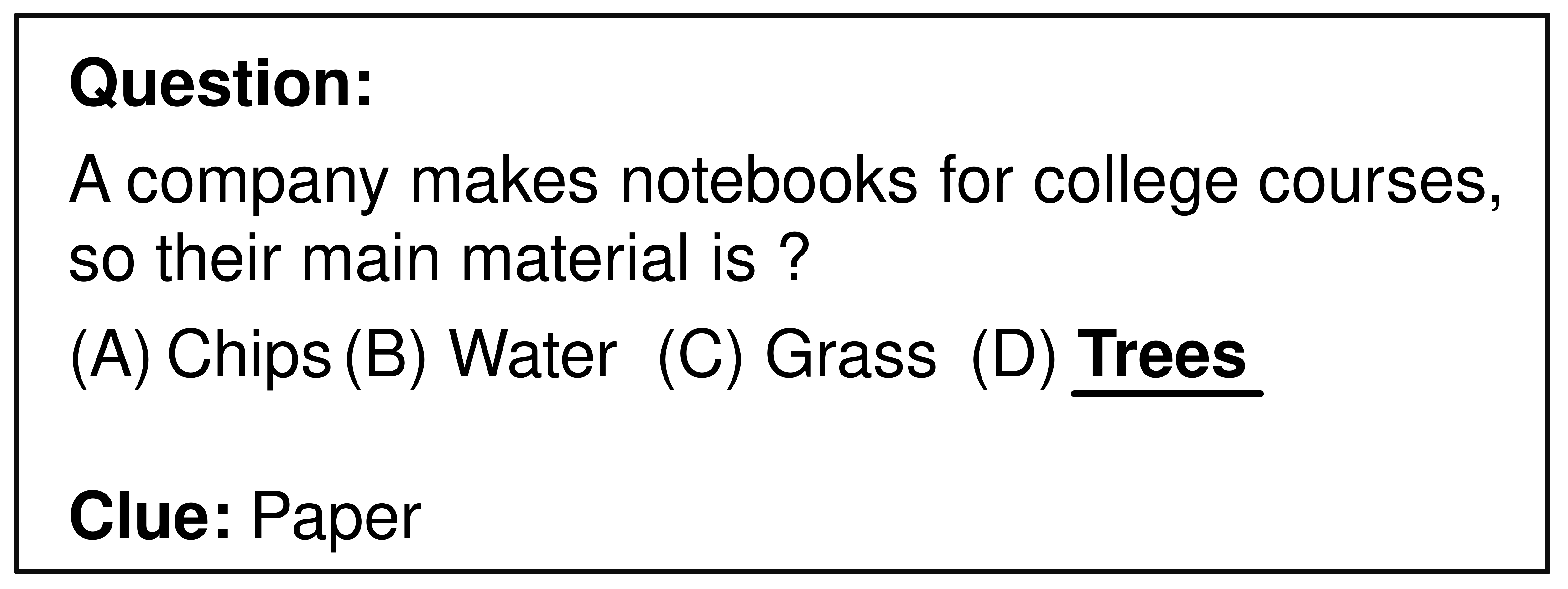}
    \caption{An example MCQA task and a generated clue. Bold underline indicates the correct answer.}
    \label{fig:example}
\end{figure}


\paragraph{Our Contribution}
Our idea is inspired by human behavior. When reading a question, humans are sometimes triggered to associate the question with their background knowledge to form some \emph{clues} even before reading the options. For simple questions, a clue may be exactly the correct answer, while for complex questions, clues may play an auxiliary role to help humans connect the question with the correct answer. For example, for the question shown in Figure~\ref{fig:example}, the clue ``paper'' forms an intermediate concept between ``notebook'' in the question and ``tree'' in the correct answer.

With this idea, we propose to employ a pre-trained encoder-decoder model to \emph{generate} a clue from the question by exploiting its underlying knowledge, without seeing and being strictly confined to the options as in the text-to-text framework. The clue representation is then leveraged by an encoder-based model to read the options and make prediction. We refer to this generation-enhanced MCQA model as \textbf{GenMC}. It significantly outperforms comparable models, in particular, text-to-text models, on five MCQA datasets.

\paragraph{Outline}
We discuss related work in Section~\ref{sec:rw}, introduce GenMC in Section~\ref{sec:model}, describe the experimental setup in Section~\ref{sec:experiments}, report the results in Section~\ref{sec:results}, and conclude in Section~\ref{sec:conclusion}.

\paragraph{Code}
Our code is available on GitHub\footnote{\url{https://github.com/nju-websoft/GenMC}} under the Apache Licence~2.0.
\section{Related Work}
\label{sec:rw}

\subsection{Text-to-Text Paradigm for MCQA}
\begin{figure}
    \subfloat[Text-to-Text]{\includegraphics[width=\columnwidth]{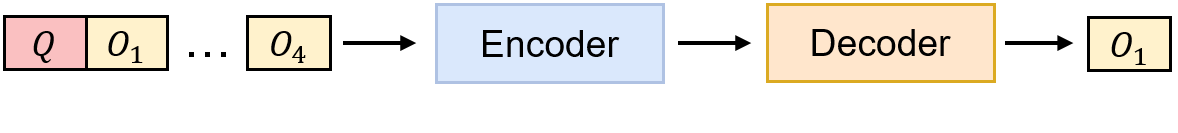}\label{fig:t2t}}\\
    \subfloat[Encoder-Only]{\includegraphics[width=\columnwidth]{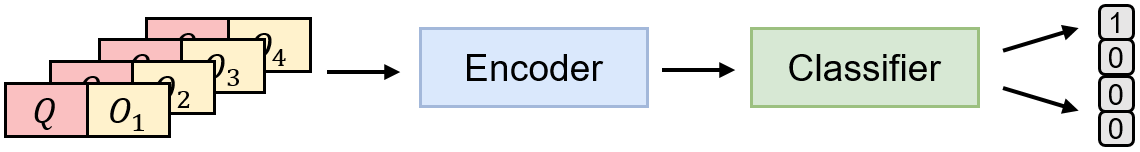}\label{fig:enc}}
    \caption{Paradigms for MCQA.}
    \label{fig:paradigm}
\end{figure}

Recently, the text-to-text paradigm has achieved breakthrough results on many NLP tasks~\cite{T5, BART}. As illustrated in Figure~\ref{fig:t2t}, adopting this paradigm for MCQA, the question~$Q$ and all the options $\{O_1,O_2,O_3,O_4\}$ are spliced into a text as input, and the correct answer~$O_1$ is used as the generation target. One benefit is that extensive training data can be shared across different tasks. Using such a framework, UnifiedQA~\cite{UnifiedQA} integrates 20 QA datasets into a unified format for training, and achieves state-of-the-art results on multiple MCQA datasets. Similarly, CALM~\cite{CALM} 
learns concept-centric knowledge from text for commonsense QA.

However, it might be debatable whether it is appropriate to train a classification task via a generation target. \citet{EncT5} point out that the decoder layers of T5 are under-utilized when fine-tuning on classification and regression tasks. Therefore, they propose a method to reduce the number of T5 parameters to improve efficiency without reducing accuracy. By contrast, we address this issue from a different perspective of how to exploit the NLG capability of pre-trained encoder-decoder models for MCQA to improve accuracy.

Some other works propose new pre-trained models for unified generation and classification tasks by designing universal encoders and task-specific decoders~\cite{CPT, ERNIE3}. They are orthogonal to our work as we leverage existing pre-trained encoder-decoder models instead of pre-training new models at an additional cost.

\subsection{Encoder-Only Paradigm for MCQA}

Benefiting from the powerful NLU capabilities of BERT-style PLMs~\cite{BERT,RoBERTa, ALBert}, the encoder-only paradigm has been popular for MCQA. As illustrated in Figure~\ref{fig:enc}, in this paradigm, the question~$Q$ and each option in $\{O_1,O_2,O_3,O_4\}$ are interacted to calculate a score, and the option with the highest score is chosen as the answer. Building on this, some works study how to design better attention-based models to identify evidence~\cite{CSA, DCMN,DUMA}. Other efforts mimic human behavior of reading evidence and answering questions~\cite{OCN,MMN,MRC_Strategy}. There, evidence is derived from the given passage or retrieved from external corpora. By contrast, we aim at exporting clues from pre-trained models without resorting to extra sources.

\begin{figure*}
    \centering
    \includegraphics[width=\textwidth]{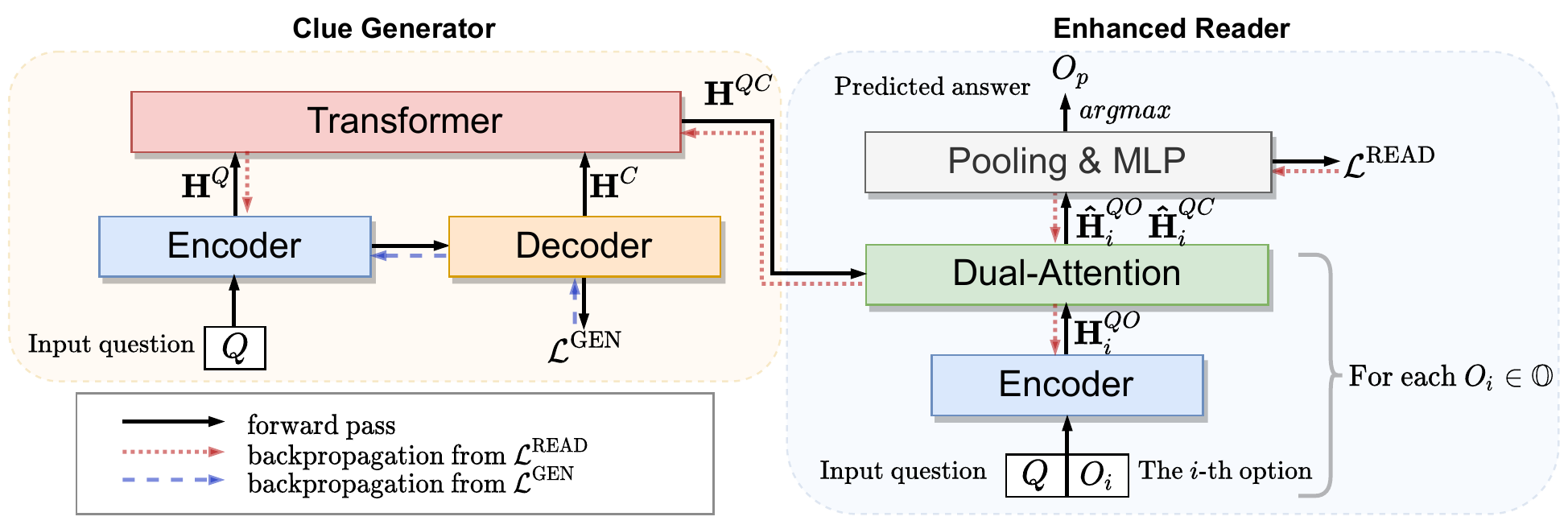}
    \caption{Architecture of GenMC. To make the prediction~$O_p \in \mathbb{O}$, the clue generator first takes $Q$ as input and outputs a clue representation~$\mathbf{H}^{QC}$ which is indicative of the correct answer. The enhanced reader then relies on the generated clue representation to better attend to options from $\mathbb{O}$ and makes the final prediction. The whole model is trained in an end-to-end manner with both the generation loss~$\mathcal{L}^\text{GEN}$ and the classification loss~$\mathcal{L}^\text{READ}$.}
    \label{fig:model}
\end{figure*}

\subsection{Knowledge in PLMs}

Recently, PLMs have been used as knowledge bases~\cite{LM_AS_KB}, and the knowledge in parameters can be exported via methods such as Prompt~\cite{What_LM_Know,AutoPrompt}. Exploiting the knowledge in PLMs for QA tasks has come into play in many forms including question expansion~\cite{GAR} and question generation~\cite{QG-Self-Talk}.

There is also research on MCQA trying to exporting knowledge from PLMs before answering. \citet{CoS-E} propose CAGE as a framework for generating explanations for commonsense QA. However, CAGE relies on explanations annotated by humans, which are not available in many real scenarios and datasets. \citet{ExplainQA} propose a joint generator-classifier model where the generator produces a human-readable textual hypothesis. Although it somewhat improves the explainability of MCQA, in terms of accuracy of MCQA there is little advancement.
CEGI~\cite{CEGI} is probably the most similar work to ours. It first uses a generative model to generate evidence, and then uses a reading model to incorporate the evidence and predict the answer, both using answer supervision. However, the generative model and the reading model are separate steps in a pipeline and are connected only via the evidence text. Such token-level interaction can lead to significant losses in accuracy as we will see in our experiments, where our representation-level interaction exhibits better performance.
\section{GenMC Model}
\label{sec:model}

In MCQA,
a question~$Q$ is given together with a set of $n$ options $\mathbb{O}=\{O_1, \ldots, O_n\}$ with exactly one option being the correct answer. The key to finding the correct answer is to capture and deeply understand the connection between~$Q$ and each $O_i \in \mathbb{O}$, which oftentimes is beyond the lexical level and requires a non-trivial entailment process.
We follow the trend of building on a pre-trained encoder-decoder model and use the encoder to jointly encode~$Q$ and each $O_i$. However, previous works directly use the decoder to generate an option in~$\mathbb{O}$, i.e.,~using the decoder as a classifier, which may have under-exploited the model's NLG capability~\cite{EncT5}.
Moreover, a simple joint encoding of~$Q$ and each $O_i$ can only enable lexical-level reasoning~\cite{zellers-etal-2019-hellaswag} which is insufficient for MCQA tasks.

Our proposed model GenMC overcomes these limitations.
Building on a pre-trained encoder-decoder model, GenMC firstly generates a clue which is indicative of the correct answer, thereby exploiting the NLG capability and underlying knowledge of the pre-trained encoder-decoder model. Then GenMC employs the generated clue representation as intermediate knowledge connecting the question and the correct answer to interact with and enhance a reader for solving MCQA. Our model design mimics how humans solve an MCQA task, i.e., after reading a question, humans may firstly associate it with some of their background knowledge (i.e., looking for clues) that helps them to later identify the correct answer.

The overall architecture of GenMC is shown in Figure~\ref{fig:model}. The clue generator (Section~\ref{subsec:clue}) first generates a clue representation only given~$Q$. Then the enhanced reader (Section~\ref{subsec:read}) uses the generated clue to augment question-option understanding.



\subsection{Clue Generator}
\label{subsec:clue}

The clue generator takes the question $Q$ as input and autoregressively outputs a clue $C={c_1, \dots, c_{|C|}}$ using a pre-trained encoder-decoder model.\footnote{For efficiency, we decode the clue greedily without performing beam search.}
Note that not the clue text~$C$ but its representation~$\mathbf{H}^C$ will be used in our model, although one could output~$C$ as evidence for explainability.

Specifically, we obtain the question representation $\mathbf{H}^Q\in\mathbb{R}^{d\times |Q|}$ and the clue representation $\mathbf{H}^C\in\mathbb{R}^{d\times |C|}$ from the last layer of the encoder and of the decoder, respectively, where $d$~denotes the representation dimension. $\mathbf{H}^C_j$, denoting the representation of the $j$-th token $c_j \in C$, is computed as follows:
\begin{equation}
\label{eq:dec}
\small
      \mathbf{p}^C_j, \mathbf{H}^C_j = \dec( c_{j-1}, \mathbf{H}_{< j}^C, \mathbf{H}^Q ) \,,
\end{equation}
where \dec$(\cdot,\cdot,\cdot)$ takes the last token $c_{j-1}$, the representation for the decoding history $\mathbf{H}_{< j}^C$, and $\mathbf{H}^Q$ as input, and outputs the hidden state $\mathbf{H}_j^C$ together with the probability distribution~$\mathbf{p}^C_j$ over the decoding vocabulary at the $j$-th step.







To encourage the tokens in~$C$ to thoroughly interact with each other and with~$Q$, we strengthen the clue representation by passing it to a transformer layer~\cite{transformer} and obtain $\mathbf{H}^{QC}$:

\begin{equation}
\small
      \mathbf{H}^{QC} = \transformer([\mathbf{H}^Q;\mathbf{H}^C] ) \,,
\end{equation}
where $[\cdot;\cdot]$ denotes concatenation. $\mathbf{H}^{QC}$ carries the information of $C$ which can be helpful to better understand and answer~$Q$.

\subsection{Enhanced Reader}
\label{subsec:read}

Previous works often directly model the relevance of each $O_i \in \mathbb{O}$ to~$Q$ via joint encoding using a pre-trained encoder, which largely performs superficial lexical reasoning~\cite{zellers-etal-2019-hellaswag}. By contrast, we use the previously generated clue representation to enhance our reader for a deeper understanding of each question-option pair.

Specifically, we first concatenate~$Q$ and each~$O_i$ independently\footnote{A delimiter "$\backslash n$" is inserted between $Q$ and each $O_i$.} and feed the concatenated input into the pre-trained encoder (which is shared with our clue generator) to obtain $O_i$'s contextualized representation $\mathbf{H}_i^{QO}$, which constitutes a column of $\mathbf{H}^{QO}\in \mathbb{R}^{d\times n}$ where $n=|\mathbb{O}|$.


Next, based on the clue representation $\mathbf{H}^{QC}$, our model intensively reads each question-option pair and obtains the matching signal between the clue and the option. Specifically, inspired by~\citet{JEEVES}, we first use dual-attention~\cite{RikiNet} to fuse information from $\mathbf{H}^{QO}_i$ to $\mathbf{H}^{QC}$
and from $\mathbf{H}^{QC}$ to $\mathbf{H}^{QO}_i$. Then we perform max-pooling to aggregate the matching features:
\begin{align}
\small
\begin{split}
    (\hat{\mathbf{H}}_{i}^{QO}, \hat{\mathbf{H}}_{i}^{QC}) & = \text{DualAttention}(\mathbf{H}_i^{QO}, \mathbf{H}^{QC}) \,,\\
    \mathbf{f}^{QO}_{i} & = \text{Max-Pooling}(\hat{\mathbf{H}}_{i}^{QO}) \,,\\
    \mathbf{f}^{QC}_{i} & = \text{Max-Pooling}(\hat{\mathbf{H}}_{i}^{QC}) \,.
\end{split}
\end{align}

To obtain the final score~$s_i$ for each~$O_i$, we concatenate the dual matching features $\mathbf{f}^{QO}_{i}$ and $\mathbf{f}^{QC}_{i}$ and feed them into a two-layer multi-layer perceptron~(MLP):
\begin{equation}
\small
      s_{i} = \text{Linear}(\text{ReLU}(\text{Linear}( [\mathbf{f}^{QO}_{i}; \mathbf{f}^{QC}_{i}])))\,.\\
\end{equation}
We select the option with the highest score as the predicted answer, denoted as~$O_p$.

\subsection{Training Objective}
We jointly train the clue generator and the enhanced reader in an end-to-end fashion with a combined loss:
\begin{equation}
\label{eq:loss}
\small
    \mathcal{L} = \mathcal{L}^\text{GEN} + \mathcal{L}^\text{READ} \,.
\end{equation}

\paragraph{Generator Loss}
For $\mathcal{L}^\text{GEN}$, assuming that $O_t \in \mathbb{O}$ is the correct answer containing $m$~tokens $a_1, \dots, a_m$, we first use $O_t$ as the target to calculate our clue generator
loss with teacher forcing:
\begin{align}
\small
\begin{split}
    \mathbf{p}^{O_t}_j, \mathbf{H}^{O_t}_j & = \dec( a_{j-1}, \mathbf{H}_{< j}^{O_t}, \mathbf{H}^Q ) \,,\\
    \mathcal{L}^\text{GEN} & = - \frac{1}{m}\sum_{j=1}^{m} \log  \mathbf{p}^{O_t}_{j, a_j} \,,
\end{split}
\end{align}
\noindent where $\mathbf{p}^{O_t}_j$~denotes the probability distribution over the decoding vocabulary at the $j$-th step, and $\mathbf{p}^{O_t}_{j, a_j}$~is the probability of token~$a_j$.

\begin{table*}[t!]
\centering
\small
\begin{tabular}{lrrrrrr}
    \hline
    & \tabincell{c}{Train set \\ size}  & \tabincell{c}{Dev set \\ size} & \tabincell{c}{Test set \\ size} & \tabincell{c}{Option \\ number}
    & \tabincell{c}{Question \\ average length} & \tabincell{c}{Option \\ average length} \\ 
    \hline
    CSQA & 8,500 & 1,241 & 1,221 & 5 & 13.38 & 1.52 \\
    OBQA & 4,957 & 500 & 500 & 4 & 10.65 & 2.85 \\
    ARC-Easy & 2,241 & 567 & 2,365 & 4 & 19.36 & 3.73\\
    ARC-Challenge & 1,117 & 295 & 1,165 & 4 & 22.30 & 4.93 \\
    QASC & 7,320 & 814 & 926 & 8 & 8.12 & 1.64\\
    \hline
\end{tabular}
\caption{Dataset statistics. For CSQA and QASC, their official dev sets are used as our test sets, and our dev sets are in-house split from their official training sets.}
\label{table:datasetstatistics}
\end{table*}

\paragraph{Reader Loss}
For $\mathcal{L}^\text{READ}$, we simply calculate a cross-entropy loss given the correct answer $O_t \in \mathbb{O}$ as follows:
\begin{equation}
\small
    \mathcal{L}^\text{READ} = -\log \frac{\exp(s_t)}{\sum_{i=1}^n\exp(s_i)} \,.
\end{equation}
Note that we update the encoder using the joint loss $\mathcal{L}$, while we do not allow $\mathcal{L}^\text{READ}$ to be backpropagated to the decoder part to reduce the memory consumption.

The above training objective exploits the double properties of the correct answer~$O_t$ in MCQA: as a text and as an index. We use~$O_t$ as a text to supervise our clue generator, and as an index (i.e., classification label) to supervise our enhanced reader. Such usage is more natural than the text-to-text paradigm~\cite{UnifiedQA,CALM}, thus having the potential to outperform.
\section{Experimental Setup}
\label{sec:experiments}

\subsection{Data}

We conducted experiments on five popular MCQA datasets spanning from commonsense questions to scientific questions. The former requires commonsense knowledge and reasoning, and the latter requires inference over scientific facts.

\paragraph{Datasets}
CSQA~\cite{CSQA} and OBQA~\cite{OBQA} are two commonsense MCQA datasets created by crowd workers based on commonsense facts. Each question is given with 5~options in CSQA and 4~options in OBQA. ARC-Easy and ARC-Challenge, denoting two disjointed subsets of ARC~\cite{ARC}, contain natural grade-school science questions with 4 options, where ARC-Challenge comprises difficult questions which require more advanced reasoning. QASC~\cite{QASC} is collected from elementary and middle school level science with 8 options for each question.

\paragraph{Train-Dev-Test Split}
For OBQA, ARC-Easy, and ARC-Challenge we used their official train, dev, and test sets. For CSQA and QASC, since the correct answers in the official test set are not public, we took their official dev set as our test set for experiments and randomly held out an in-house dev set from the training set. The dataset statistics are shown in Table~\ref{table:datasetstatistics}.

\paragraph{External Knowledge}
For all these datasets, our experiments did not rely on any provided documents or external corpora; a question was solely provided with its options to form the input. It means that pre-trained models were used as the primary source of knowledge in the experiments.


\subsection{Implementation Details}
\label{sec:experiments-details}

We used two popular encoder-decoder models as a basis, BART~\cite{BART} and T5~\cite{T5}. For each model, we experimented with its \textsc{Base} and \textsc{Large} versions.

We used PyTorch 1.7. We used the Adam optimizer and set $\text{warmup fraction} = 0.1$, $\text{weight decay} = 0.01$, $\text{maximum source length}=64$, $\text{maximum target length} = 32$, $\text{epoch}=30$, and early stop training when there was no better result on the dev set after 5 epochs. For each model, we searched for the best learning rate from $\{1e-4, 5e-5, 1e-5\}$, and for the best batch size out of $\{8, 64\}$.

Because neural models are known to be sensitive to different random seeds, especially when the training set is small, we performed multiple experiments for all models with different random seeds, and reported the mean and standard deviation. For CSQA, OBQA, ARC-Easy, and QASC, we used three random seeds $\{1, 10, 20\}$. For the smallest dataset ARC-Challenge, we used five random seeds $\{1, 10, 20, 30, 40\}$.

All the experiments were performed on a GeForce RTX 3090 with 24G memory.

\subsection{Evaluation Metric}

For each model, we reported its proportion of correctly answered questions in each dataset.
\section{Experimental Results}
\label{sec:results}

\begin{table*}[t!]
\centering
\resizebox{\textwidth}{!}{$
\begin{tabular}{lccccccccccc}
\hline
& \multicolumn{2}{c}{CSQA} & \multicolumn{2}{c}{OBQA} & \multicolumn{2}{c}{ARC-Easy} & \multicolumn{2}{c}{ARC-Challenge} & \multicolumn{2}{c}{QASC} \\
\cline{2-11}
& dev & test & dev & test & dev & test  & dev & test  & dev & test \\
\hline
BART$_\textsc{Base}$ \\
\quad Text2Text$_\text{vanilla}$ & 51.62~{\small($\pm$0.04)} & 53.26~{\small($\pm$0.57)} & 54.93~{\small($\pm$0.83)} & 52.73~{\small($\pm$1.00)} & 51.55~{\small($\pm$1.38)} & 50.51~{\small($\pm$1.82)} & 30.05~{\small($\pm$1.25)} & 24.95~{\small($\pm$1.10)} & 46.72~{\small($\pm$1.21)} & 26.78~{\small($\pm$1.21)}   \\

\quad Text2Text$_\text{enc}$ & 50.63~{\small($\pm$0.66)} & 52.22~{\small($\pm$1.64)} & 55.87~{\small($\pm$1.10)} &  51.00~{\small($\pm$1.83)} & 49.03~{\small($\pm$1.86)} & 49.94~{\small($\pm$1.49)} & 32.32~{\small($\pm$4.87)} & 26.24~{\small($\pm$2.01)} & 48.08~{\small($\pm$1.35)} & 17.06~{\small($\pm$0.39)} \\

\quad GenMC & \textbf{54.82}~{\small($\pm$0.61)} & \textbf{56.40}~{\small($\pm$0.61)} & \textbf{58.53}~{\small($\pm$0.31)} & \textbf{57.53}~{\small($\pm$2.91)} & \textbf{59.38}~{\small($\pm$1.60)} & \textbf{56.80}~{\small($\pm$0.28)} & \textbf{38.64}~{\small($\pm$0.90)} & \textbf{33.82}~{\small($\pm$1.66)} & \textbf{57.70}~{\small($\pm$0.43)} & \textbf{35.96}~{\small($\pm$1.70)} \\

\hline
T5$_\textsc{Base}$ \\
\quad Text2Text$_\text{vanilla}$ & 57.59~{\small($\pm$0.81)} & 60.93~{\small($\pm$0.73)} & 59.53~{\small($\pm$0.81)} &  57.53~{\small($\pm$0.70)} & 52.20~{\small($\pm$0.31)} & 51.75~{\small($\pm$0.89)} & 29.38~{\small($\pm$2.63)} & 23.69~{\small($\pm$2.47)} & 54.55~{\small($\pm$1.01)} & 37.94~{\small($\pm$1.47)}   \\
\quad Text2Text$_\text{enc}$ & 58.96~{\small($\pm$1.21)} & 59.49~{\small($\pm$1.41)} & 60.67~{\small($\pm$2.86)} &  57.07~{\small($\pm$3.03)} & 56.55~{\small($\pm$1.17)} & 52.92~{\small($\pm$0.29)} & 29.49~{\small($\pm$5.13)} & 26.09~{\small($\pm$0.23)} & 56.84~{\small($\pm$0.84)} & 39.60~{\small($\pm$2.38)}   \\
\quad GenMC & \textbf{60.65}~{\small($\pm$0.47)} & \textbf{63.45}~{\small($\pm$0.29)} & \textbf{62.07}~{\small($\pm$1.01)} & \textbf{61.67}~{\small($\pm$0.58)} & \textbf{62.38}~{\small($\pm$0.67)} & \textbf{58.82}~{\small($\pm$0.37)} & \textbf{43.62}~{\small($\pm$0.52)} & \textbf{39.00}~{\small($\pm$0.30)} & \textbf{58.93}~{\small($\pm$1.76)} & \textbf{41.72}~{\small($\pm$1.18)}  \\
\hline

BART$_\textsc{Large}$ \\
\quad Text2Text$_\text{vanilla}$ & 65.58~{\small($\pm$2.72)} & 66.91~{\small($\pm$2.14)} & 62.66~{\small($\pm$1.18)} &  61.46~{\small($\pm$1.74)} & 63.49~{\small($\pm$1.89)} & 62.81~{\small($\pm$2.15)} & 29.94~{\small($\pm$2.32)} & 28.55~{\small($\pm$4.97)} & 64.57~{\small($\pm$2.21)} & 47.80~{\small($\pm$2.22)}  \\

\quad Text2Text$_\text{enc}$ & 65.00~{\small($\pm$0.66)} & 67.35~{\small($\pm$0.90)} & 63.80~{\small($\pm$1.44)} &  62.47~{\small($\pm$1.53)} & 68.20~{\small($\pm$2.04)} & 65.33~{\small($\pm$1.74)} & 35.37~{\small($\pm$6.07)} & 31.13~{\small($\pm$5.86)} & 65.07~{\small($\pm$0.94)} & 47.19~{\small($\pm$0.71)}  \\

\quad GenMC & \textbf{69.57}~{\small($\pm$0.89)} & \textbf{72.26}~{\small($\pm$0.70)} & \textbf{68.93}~{\small($\pm$1.17)} & \textbf{68.07}~{\small($\pm$1.70)} & \textbf{72.43}~{\small($\pm$0.54)} & \textbf{68.68}~{\small($\pm$0.34)} & \textbf{48.93}~{\small($\pm$0.98)} & \textbf{45.52}~{\small($\pm$1.54)} & \textbf{68.39}~{\small($\pm$0.68)} & \textbf{55.90}~{\small($\pm$0.92)}  \\
\hline

T5$_\textsc{Large}$ \\
\quad Text2Text$_\text{vanilla}$ & 67.53~{\small($\pm$0.43)} & 70.63~{\small($\pm$0.74)} & 66.80~{\small($\pm$0.87)} &  63.53~{\small($\pm$1.10)} & 65.61~{\small($\pm$0.18)} & 62.55~{\small($\pm$0.54)} & 43.05~{\small($\pm$1.69)} & 42.83~{\small($\pm$2.00)} & 64.13~{\small($\pm$1.47)} & 57.74~{\small($\pm$0.82)}  \\

\quad Text2Text$_\text{enc}$ & 68.41~{\small($\pm$0.73)} & 70.30~{\small($\pm$0.82)} & 65.93~{\small($\pm$1.03)} &  63.67~{\small($\pm$0.46)} & 69.61~{\small($\pm$0.20)} & 66.65~{\small($\pm$0.34)} & 30.73~{\small($\pm$3.15)} & 28.76~{\small($\pm$4.85)} & 65.27~{\small($\pm$1.55)} & 55.65~{\small($\pm$0.45)}   \\

\quad GenMC & \textbf{71.10}~{\small($\pm$0.41)} & \textbf{72.67}~{\small($\pm$1.02)} & \textbf{71.60}~{\small($\pm$0.92)} & \textbf{66.87}~{\small($\pm$1.33)} & \textbf{72.49}~{\small($\pm$0.77)} & \textbf{69.01}~{\small($\pm$1.97)} & \textbf{49.83}~{\small($\pm$2.06}) & \textbf{47.41}~{\small($\pm$2.00)} & \textbf{67.61}~{\small($\pm$1.14)} & \textbf{58.06}~{\small($\pm$0.92)}  \\
\hline
\end{tabular}
$}

\caption{Comparison with text-to-text models.}
\label{table:comp_t2t_model}
\end{table*}

\subsection{Main Results: Comparison with Text-to-Text Models}
\label{subsec:main_results}

To empirically evaluate GenMC in terms of whether it better exploits the potential of pre-trained encoder-decoder models for MCQA, we compare GenMC with a standard text-to-text implementation and with a variant thereof for analysis.

\subsubsection{Baselines}

\paragraph{Text2Text$_\text{vanilla}$} 
The vanilla usage of pre-trained encoder-decoders for MCQA is to reform the input and output in a way that can be directly processed by a encoder-decoder model. Specifically, following~\citet{T5}, we concatenate the input question with all candidate options, where each option is also preceded by its option ID, and then prepend the sequence with a dataset name. The concatenated sequence is fed into the encoder part to get a joint representation for the question and all options. Based on the joint representation, the decoder finally outputs an option ID. In this setting, the decoder is basically used as a classifier.


\paragraph{Text2Text$_\text{enc}$}
Similar to~\citet{EncT5}, we use only the encoder part of a pre-trained encoder-decoder model. Each option is independently paired with the question to obtain a joint representation using the encoder. Then the representation is fed into a scorer (i.e., an MLP) to obtain a matching score for each question-option pair. The model then predicts the option with the highest score. In this setting, the decoder is totally unused. Though~\citet{EncT5} find that their encoder-only model performs comparably to using the decoder as a classifier, we argue that the decoder part can further improve the performance, if being properly used.


\subsubsection{Results}

The main results (see Table~\ref{table:comp_t2t_model}) show that GenMC consistently and significantly (with p-value < 0.01) outperforms Text2Text$_\text{vanilla}$ and Text2Text$_\text{enc}$ on all datasets. For several settings, GenMC even obtains an absolute gain of over 10\%. For example, on the test set of the challenging scientific MCQA dataset ARC-Challenge, T5$_\textsc{Base}$ + GenMC improves T5$_\textsc{Base}$ + Text2Text$_\text{vanilla}$ from an accuracy of 23.69\% to 39.00\%, suggesting a relative gain of around 65\%. These results demonstrate that GenMC is a more effective usage of pre-trained encoder-decoder models than existing ones.

Moreover, we interestingly find that the decoder-free baseline Text2Text$_\text{enc}$ outperforms Text2Text$_\text{vanilla}$ on over half of the experiments. This indicates that the decoder's general language knowledge gained from pre-training is largely wasted by only using it as a classifier, which may further explain the superior performance of our model because GenMC can exploit the pre-trained decoder more effectively. In addition, all $\textsc{Large}$ models significantly outperform their $\textsc{Base}$ counterparts. This suggests that the embedded knowledge gained from pre-training is critical to MCQA tasks, strengthening our point to make full use of pre-trained encoders and decoders.


\subsection{Comparison with Other Models}


\begin{table*}[t!]
\centering
\resizebox{\textwidth}{!}{
\begin{tabular}{lccccccccccc}
\hline
& \multicolumn{2}{c}{CSQA} & \multicolumn{2}{c}{OBQA} & \multicolumn{2}{c}{ARC-Easy} & \multicolumn{2}{c}{ARC-Challenge} & \multicolumn{2}{c}{QASC} \\
\cline{2-11}
& dev & test & dev & test & dev & test  & dev & test  & dev & test \\
\hline
\textsc{Base} \\
\quad RoBERTa & 56.51~{\small($\pm$0.34)} &
58.91~{\small($\pm$0.79)} & 58.67~{\small($\pm$1.03)} & 49.67~{\small($\pm$0.76)} & 56.56~{\small($\pm$0.91)} & 52.32~{\small($\pm$0.70)} & 38.64~{\small($\pm$0.90)} & 34.85~{\small($\pm$2.20)} & 55.28~{\small($\pm$0.12)} & 34.38~{\small($\pm$1.72)}  \\  
\quad ALBERT & 
53.16~\small($\pm$0.58) & 53.95~\small($\pm$0.49) & 54.53~\small($\pm$1.10) &  49.20~\small($\pm$2.27) & 48.32~\small($\pm$0.88) & 45.84~\small($\pm$1.94) & 34.80~\small($\pm$1.53) & 30.21~\small($\pm$1.74) & 40.99~\small($\pm$(1.78) & 24.55~\small($\pm$1.23)  \\
\quad UnifiedQA$_\text{T5}$~$\ast$ & - & 45.00~{\small($\pm$0.00)} & - &  59.00~{\small($\pm$0.00)} & - & 53.00~{\small($\pm$0.00)} & - & 42.40~{\small($\pm$0.00)} & - & 25.80~{\small($\pm$0.00)}  \\ 
\quad UnifiedQA$_\text{T5}$ & 41.02~\small($\pm$0.00) & 44.80~\small($\pm$0.00) & 59.20~\small($\pm$0.00) &  59.60~\small($\pm$0.00) & 54.85~\small($\pm$0.00) & 53.66~\small($\pm$0.00) & 44.75~\small($\pm$0.00) & \textbf{42.58}~\small($\pm$0.00) & 17.94~\small($\pm$0.00) & 25.70~\small($\pm$0.00)  \\
\quad UnifiedQA$_\text{T5-FT}$ & 56.81~\small($\pm$0.49) & 62.35~\small($\pm$0.80) & 60.80~\small($\pm$0.72) &  58.47~\small($\pm$0.64) & 54.97~\small($\pm$0.20) & 53.88~\small($\pm$0.39) & \textbf{45.31}~\small($\pm$0.39) & 42.43~\small($\pm$0.47) & 55.57~\small($\pm$0.58) & \textbf{43.20}~\small($\pm$0.57)  \\
\quad GenMC$_\text{T5}$  & \textbf{60.65}~\small($\pm$0.47) & \textbf{63.45}~\small($\pm$0.29) & \textbf{62.07}~\small($\pm$1.01) &  \textbf{61.67}~\small($\pm$0.58) & \textbf{62.38}~\small($\pm$0.67) & \textbf{58.82}~\small($\pm$0.37) & 43.62~\small($\pm$0.52) & 
39.00~\small($\pm$0.30) & \textbf{58.93}~\small($\pm$1.76) & 41.72~\small($\pm$1.18)  \\
\hline
\textsc{Large} \\
\quad RoBERTa & 68.92~\small($\pm$0.76) & 71.88~\small($\pm$0.26) & 67.80~\small($\pm$1.22) &  64.47~\small($\pm$1.41) & 65.73~\small($\pm$0.80) & 62.40~\small($\pm$0.89) & 38.08~\small($\pm$1.99) & 35.97~\small($\pm$1.74) & 67.32~\small($\pm$0.58) & 50.22~\small($\pm$1.88)  \\  
\quad ALBERT & 60.62~\small($\pm$0.57) & 59.32~\small($\pm$0.91) & 54.50~\small($\pm$1.40) &  49.27~\small($\pm$0.64) & 54.03~\small($\pm$0.45) & 53.77~\small($\pm$1.81) & 33.90~\small($\pm$1.22) & 31.19~\small($\pm$3.79) & 51.11~\small($\pm$1.72) & 33.12~\small($\pm$1.24)  \\  
\quad UnifiedQA$_\text{T5}$~$\ast$ & - & 60.90~\small($\pm$0.00) & - &  68.40~\small($\pm$0.00) & - & 65.90~\small($\pm$0.00) & - & 54.40~\small($\pm$0.00) & - & 43.30~\small($\pm$0.00)  \\
\quad UnifiedQA$_\text{T5}$ & 55.28~\small($\pm$0.00) & 61.34~\small($\pm$0.00) & 70.40~\small($\pm$0.00) &  68.40~\small($\pm$0.00) & 69.31~\small($\pm$0.00) & 66.43~\small($\pm$0.00) & 56.61~\small($\pm$0.00) & 54.33~\small($\pm$0.00) & 29.24~\small($\pm$0.00) & 43.74~\small($\pm$0.00)  \\
\quad UnifiedQA$_\text{T5-FT}$ & 69.00~\small($\pm$0.51) & \textbf{73.60}~\small($\pm$0.45) & 70.53~\small($\pm$0.23) &  \textbf{68.80}~\small($\pm$0.69) & 69.72~\small($\pm$0.71) & 66.92~\small($\pm$0.85) & \textbf{56.84}~\small($\pm$0.39) & \textbf{54.42}~\small($\pm$0.15) & 66.63~\small($\pm$1.56) & \textbf{58.71}~\small($\pm$0.90)  \\
\quad GenMC$_\text{T5}$  & \textbf{71.10}~\small($\pm$0.41) & 72.67~\small($\pm$1.02) & \textbf{71.60}~\small($\pm$0.92) &  66.87~\small($\pm$1.33) & \textbf{72.49}~\small($\pm$0.77) & \textbf{69.01}~\small($\pm$1.97) & 49.83~\small($\pm$2.06) & 47.41~\small($\pm$2.00) & \textbf{67.61}~\small($\pm$1.14) & 58.06~\small($\pm$0.92)  \\
\hline
\end{tabular}
}
\caption{Comparison with other models. (* indicates the results reported by~\citet{UnifiedQA}.)}
\label{table:sota}
\end{table*}

\begin{table*}[t!]
\centering
\resizebox{\textwidth}{!}{$
\begin{tabular}{lccccccccccc}
\hline
& \multicolumn{2}{c}{CSQA} & \multicolumn{2}{c}{OBQA} & \multicolumn{2}{c}{ARC-Easy} & \multicolumn{2}{c}{ARC-Challenge} & \multicolumn{2}{c}{QASC} \\
\cline{2-11}
& dev & test & dev & test & dev & test  & dev & test  & dev & test \\
\hline
\textsc{Base} \\

\quad UnifiedQA$_\text{T5-FT}$ & 
56.81~\small($\pm$0.49) & 62.35~\small($\pm$0.80) & 60.80~\small($\pm$0.72) &  58.47~\small($\pm$0.64) & 54.97~\small($\pm$0.20) & 53.88~\small($\pm$0.39) & 45.31~\small($\pm$0.39) & 42.43~\small($\pm$0.47) & 55.57~\small($\pm$0.58) & 43.20~\small($\pm$0.57)  \\

\quad GenMC$_\text{T5-U}$  & \textbf{61.24}~{\small($\pm$0.45)} & \textbf{63.45}~{\small($\pm$0.76)} & \textbf{62.33}~{\small($\pm$0.81)} &  \textbf{59.20}~{\small($\pm$1.91)} & \textbf{61.73}~{\small($\pm$0.35)} & \textbf{59.35}~{\small($\pm$0.43)} & \textbf{45.54}~{\small($\pm$0.20)} & \textbf{43.98}~{\small($\pm$0.36)} & \textbf{60.16}~{\small($\pm$0.07)} & \textbf{45.43}~{\small($\pm$0.87)}  \\

\hline
\textsc{Large} \\
\quad UnifiedQA$_\text{T5-FT}$ & 69.00~\small($\pm$0.51) & \textbf{73.60}~\small($\pm$0.45) & 70.53~\small($\pm$0.23) &  68.80~\small($\pm$0.69) & 69.72~\small($\pm$0.71) & 66.92~\small($\pm$0.85) & 56.84~\small($\pm$0.39) & 54.42~\small($\pm$0.15) & 66.63~\small($\pm$1.56) & 58.71~\small($\pm$0.90)  \\ 

\quad GenMC$_\text{T5-U}$  & \textbf{71.58}~{\small($\pm$0.25)} & 72.26~{\small($\pm$0.31)} & \textbf{71.67}~{\small($\pm$0.46)} &  \textbf{69.00}~{\small($\pm$0.69)} & \textbf{73.90}~{\small($\pm$0.47)} & \textbf{72.87}~{\small($\pm$0.50)} & \textbf{59.55}~{\small($\pm$1.09)} & \textbf{55.97}~{\small($\pm$0.62)} & \textbf{68.55}~{\small($\pm$0.81)} & \textbf{58.75}~{\small($\pm$0.56)}  \\  
\hline
\end{tabular}
$}
\caption{Comparison with UnifiedQA after unifying training sets.}
\label{table:unified}
\end{table*}

\subsubsection{Baselines}

\paragraph{UnifiedQA}
Existing methods that rely on external documents or corpora have achieved state-of-the-art performance on several MCQA datasets. However, to enable a fair comparison, we only compare with models that adopt the same setting as ours, where a question and its options are the only input to the model. Among these models, UnifiedQA~\cite{UnifiedQA} is the current best model. While UnifiedQA reports the best score using its T5-11B version, since for T5 we experiment with its \textsc{Base} and \textsc{Large} versions, we only report and compare under T5$_\textsc{Base}$ and T5$_\textsc{Large}$. Note that instead of training on each dataset separately, UnifiedQA converts a line of popular QA datasets with four formats (e.g., retrieval-based QA, MCQA) into a unified format, and trains a single model over all training data, while GenMC only uses each dataset's own training data.

\paragraph{RoBERTa and ALBERT}
In addition, we compare with two encoder-only models, RoBERTa~\cite{RoBERTa} and ALBERT~\cite{ALBert}, which have served as the basis of many MCQA models.

All models are of comparable model size to ours.

\begin{table*}[t!]
\centering
\resizebox{\textwidth}{!}{$
\begin{tabular}{lccccccccccc}
\hline
& \multicolumn{2}{c}{CSQA} & \multicolumn{2}{c}{OBQA} & \multicolumn{2}{c}{ARC-Easy} & \multicolumn{2}{c}{ARC-Challenge} & \multicolumn{2}{c}{QASC} \\
\cline{2-11}
& dev & test & dev & test & dev & test  & dev & test  & dev & test \\
\hline
BART$_\textsc{Base}$ \\
\quad GenMC & \textbf{54.82}~{\small($\pm$0.61)} & \textbf{56.40}~{\small($\pm$0.61)} & \textbf{58.53}~{\small($\pm$0.31)} & \textbf{57.53}~{\small($\pm$2.91)} & \textbf{59.38}~{\small($\pm$1.60)} & \textbf{56.80}~{\small($\pm$0.28)} & 38.64~{\small($\pm$0.90)} & \textbf{33.82}~{\small($\pm$1.66)} & \textbf{57.70}~{\small($\pm$0.43)} & \textbf{35.96}~{\small($\pm$1.70)} \\
\quad Weak Clue & 53.96~{\small($\pm$1.01)} & 54.35~{\small($\pm$1.97)} & 55.53~{\small($\pm$1.27)} &  54.27~{\small($\pm$0.92)} & 57.20~{\small($\pm$1.80)} & 55.42~{\small($\pm$1.26)} & \textbf{39.89}~{\small($\pm$0.20)} & 32.62~{\small($\pm$0.31)} & 54.05~{\small($\pm$0.21)} & 25.99~{\small($\pm$0.82)}  \\  
\quad Token Clue & 45.53~{\small($\pm$1.28)}  & 46.41~{\small($\pm$1.79)}   & 54.07~{\small($\pm$1.72)}  & 52.93~{\small($\pm$1.10)}  & 48.97~{\small($\pm$0.91)}  & 48.87~{\small($\pm$1.29)}  & 31.19~{\small($\pm$0.59)}  & 27.64~{\small($\pm$0.69)}  & 49.06~{\small($\pm$0.39)}  & 21.31~{\small($\pm$1.03)}  \\  
\hline

T5$_\textsc{Base}$ \\
\quad GenMC & \textbf{60.65}~{\small($\pm$0.47)} & \textbf{63.45}~{\small($\pm$0.29)} & \textbf{62.07}~{\small($\pm$1.01)} & \textbf{61.67}~{\small($\pm$0.58)} & \textbf{62.38}~{\small($\pm$0.67)} & \textbf{58.82}~{\small($\pm$0.37)} & \textbf{43.62}~{\small($\pm$0.52)} & \textbf{39.00}~{\small($\pm$0.30)} & \textbf{58.93}~{\small($\pm$1.76)} & \textbf{41.72}~{\small($\pm$1.18)}  \\
\quad Weak Clue & 58.80~{\small($\pm$0.70)} & 60.88~{\small($\pm$1.89)} & 61.47~{\small($\pm$0.95)} &  59.73~{\small($\pm$0.90)} & 58.97~{\small($\pm$0.54)} & 57.10~{\small($\pm$0.72)} & 42.26~{\small($\pm$2.21)} & 37.54~{\small($\pm$0.64)} & 57.37~{\small($\pm$1.40)} & 36.29~{\small($\pm$1.66)}  \\  
\quad Token Clue & 50.55~{\small($\pm$0.44)}  & 48.79~{\small($\pm$0.87)} & 56.00~{\small($\pm$1.25)}  & 54.93~{\small($\pm$1.63)}  & 46.50~{\small($\pm$0.83)}  & 46.65~{\small($\pm$0.54)}  & 32.66~{\small($\pm$0.20)}  & 26.01~{\small($\pm$1.28)}  & 43.69~{\small($\pm$1.52)}  & 27.50~{\small($\pm$1.56)}  \\  
\hline

BART$_\textsc{Large}$ \\
\quad GenMC & \textbf{69.57}~{\small($\pm$0.89)} & \textbf{72.26}~{\small($\pm$0.70)} & \textbf{68.93}~{\small($\pm$1.17)} & \textbf{68.07}~{\small($\pm$1.70)} & \textbf{72.43}~{\small($\pm$0.54)} & \textbf{68.68}~{\small($\pm$0.34)} & \textbf{48.93}~{\small($\pm$0.98)} & \textbf{45.52}~{\small($\pm$1.54)} & \textbf{68.39}~{\small($\pm$0.68)} & \textbf{55.90}~{\small($\pm$0.92)}  \\
\quad Weak Clue & 67.28~{\small($\pm$2.39)} & 69.64~{\small($\pm$2.76)} & 66.20~{\small($\pm$0.53)} &  64.47~{\small($\pm$1.40)} & 70.66~{\small($\pm$1.50)} & 65.71~{\small($\pm$1.47)} & 27.80~{\small($\pm$2.06)} & 24.92~{\small($\pm$2.06)} & 65.68~{\small($\pm$1.31)} & 52.02~{\small($\pm$1.44)}  \\  
\quad Token Clue & 53.85~{\small($\pm$0.47)}  & 55.23~{\small($\pm$0.62)}  & 61.20~{\small($\pm$3.14)}  & 59.20~{\small($\pm$0.69)}  & 58.02~{\small($\pm$0.98)}  & 54.22~{\small($\pm$1.27)}  & 41.81~{\small($\pm$1.19)}  & 37.60~{\small($\pm$0.90)}  & 48.65~{\small($\pm$1.23)}  & 32.47~{\small($\pm$1.11)}  \\  
\hline

T5$_\textsc{Large}$ \\
\quad GenMC & \textbf{71.10}~{\small($\pm$0.41)} & \textbf{72.67}~{\small($\pm$1.02)} & \textbf{71.60}~{\small($\pm$0.92)} & \textbf{66.87}~{\small($\pm$1.33)} & \textbf{72.49}~{\small($\pm$0.77)} & \textbf{69.01}~{\small($\pm$1.97)} & \textbf{49.83}~{\small($\pm$2.06)} & \textbf{47.41}~{\small($\pm$2.00)} & \textbf{67.61}~{\small($\pm$1.14)} & \textbf{58.06}~{\small($\pm$0.92)}  \\
\quad Weak Clue & 68.33~{\small($\pm$1.62)} & 71.66~{\small($\pm$1.28)} & 69.27~{\small($\pm$0.42)} &  65.87~{\small($\pm$0.90)} & 69.66~{\small($\pm$0.77)} & 66.24~{\small($\pm$0.79)} & 47.57~{\small($\pm$2.04)} & 46.24~{\small($\pm$1.29)} & 64.99~{\small($\pm$0.74)} & 53.35~{\small($\pm$1.35)}  \\  
\quad Token Clue & 59.47~{\small($\pm$0.08)}  & 60.74~{\small($\pm$0.29)}  & 62.80~{\small($\pm$1.44)}  & 57.73~{\small($\pm$1.10)}  & 48.85~{\small($\pm$1.62)}  & 48.36~{\small($\pm$2.15)}  & 37.97~{\small($\pm$0.90)}  & 30.50~{\small($\pm$1.46)}  & 49.22~{\small($\pm$0.62)}  & 38.77~{\small($\pm$1.74)}  \\  
\hline
\end{tabular}
$}
\caption{Influence of clues.}
\label{table:wo_clue}
\end{table*}

\subsubsection{Results}

The results in Table~\ref{table:sota} show that GenMC$_\text{T5}$  significantly (with p-value < 0.01) outperforms the two encoder-only strong baselines RoBERTa and ALBERT. More interestingly, GenMC$_\text{T5}$  also performs better than UnifiedQA$_\text{T5}$ on most datasets. Moreover, for UnifiedQA$_\text{T5-FT}$, which further fine-tunes the model on the training set of the target dataset, GenMC$_\text{T5}$  outperforms it on the test sets of CSQA, OBQA, and ARC-Easy for the base models and ARC-Easy for the large models. It also achieves comparable results on the remaining datasets. These results are impressive because UnifiedQA uses more datasets (i.e., eight different QA datasets) for training. The promising results of GenMC further reveals that our model can learn to effectively extract knowledge from pre-trained encoder-decoders with limited training data.

As a fairer comparison in Table~\ref{table:unified}, by unifying the training sets of all the five datasets, our GenMC$_\text{T5-U}$  outperforms UnifiedQA$_\text{T5-FT}$ on all datasets except for CSQA with large models.

\subsection{Ablation Study: Influence of Clues}
\label{sec:influence_clue}

Our main results in Section~\ref{subsec:main_results} have demonstrated the effectiveness of our model. To better understand its superior results and the influence of our clue generation, we compare with two variants.

\subsubsection{Variants of GenMC}

\paragraph{Weak Clue}
We train this variant only using the classification loss $\mathcal{L}^\text{READ}$, so only the encoder part is updated, while the decoder part is left untouched from pre-training. Intuitively, under this setting, the generated clue is weaker than GenMC which learns how to generate a clue with supervision.

\paragraph{Token Clue}
In this setting, we separately train a clue generator and a reader. We first collect the generated clue text~$C$ (instead of its representation) from the decoder. We then directly concatenate~$C$ with~$Q$ and~$O_i$ to compute a score for~$O_i$ using the model's encoder part stacked with an MLP layer. This variant is indeed very similar to~\citet{CEGI}, which also adopts a pipeline framework to first generate a token-level evidence and then use the evidence to expand the question.

\begin{table*}[t!]
\centering
\resizebox{\textwidth}{!}{$
\begin{tabular}[t]{@{}lrll@{}}
\hline
\multirow{2}{*}{Clue Type} & \multirow{2}{*}{Percentage} & \multicolumn{2}{c}{Example} \\ \cline{3-4} 
& & Instance & Clue \\ \hline
Irrelevant         & 23.60\% &    \begin{tabular}[c]{@{}l@{}}Which would you likely find inside a beach ball?\\ (A) cheese\quad(B) \textit{steam}\quad(C) water\quad (D) \textbf{\underline{air}}\end{tabular}       &     a squid       \\
Relevant but unhelpful  & 52.40\% &  \begin{tabular}[c]{@{}l@{}}What may have been formed by a volcano?\\ (A) \textbf{\underline{Mt. McKinley}}\quad(B) Lake Pontchartrain \quad(C) The great lakes\quad \\(D) \textit{Niagara Falls}\end{tabular}        &      a lake          \\
Helpful & 24.00\% &   \begin{tabular}[c]{@{}l@{}}Where would there be an auditorium with only a single person speaking?\\ (A) lights \quad(B) crowd \quad (C) \textbf{\underline{university campus}}\quad (D) \textit{theater}\quad (E) park\end{tabular}        &     school            \\ \hline
\end{tabular}
$}
\caption{Distribution of clue types in negative cases with examples. Bold underline indicates the correct answer, and italic indicates the predicted label.}
\label{table:negExp}
\end{table*}

\subsubsection{Results}

Table~\ref{table:wo_clue} shows that masking out generation loss leads to substantial performance drops across all datasets, demonstrating that fine-tuning the decoder with generation loss $\mathcal{L}^\text{GEN}$ helps to derive useful clues from pre-trained encoder-decoder models. We also observe that the performance of using token-level clues lags much behind GenMC. This demonstrates that naively using explicit knowledge in plain text, instead of using implicit clues from the decoder's hidden state, is inferior as it may unnecessarily bring information loss and noise.





\subsection{Error Analysis}
We analyze the clues generated by GenMC using T5$_\textsc{Large}$ with a focus on instances that are correctly predicted by the baseline in our main experiments (i.e., T5$_\textsc{Large}$ + Text2Text$_\text{vanilla}$), while our GenMC fails. The intuition is that in these \emph{negative cases}, the clues generated by GenMC may play a negative role. By studying these potentially negative clues, we can gain more insights into how GenMC fails and discuss venues for future improvement.

Specifically, we randomly sample 50 negative cases from T5$_\textsc{Large}$ + GenMC for each dataset. We show six graduate students of computer science\footnote{They are volunteers recruited from the contact author's research group. They know and agree that their annotations will be used for error analysis in a research paper.} an instance along with the generated clue, correct answer, and predicted answer. We then ask them to categorize clues into the following families:\footnote{We follow a similar definition by~\citet{QG-Self-Talk}.}

\begin{itemize}
    \item \textbf{Irrelevant:} The clue is off topic or is not understandable.
    \item \textbf{Relevant but unhelpful:} Though relevant, the clue makes a factually incorrect statement, often on the contrary of the main question, or the clue contributes relevant but insufficient knowledge for prediction, such as repetition of the question or other distractors.
    \item \textbf{Helpful:} The clue adds helpful information to answer the question.
\end{itemize}

\begin{table*}[!t]
\centering
\small
\begin{tabular}{lccccc}
    \hline
    & CSQA  & OBQA & ARC-Easy & ARC-Challenge & QASC \\ 
    \hline
    T5$_\textsc{Base}$ \\
    \quad Text2Text$_\text{vanilla}$ & 0.040~{\small($\pm$0.007)} & 0.035~{\small($\pm$0.002)} & 0.035~{\small($\pm$0.002)} & 0.039~{\small($\pm$0.004)} & 0.035~{\small($\pm$0.002)} \\
    
    \quad UnifiedQA 
    & 0.059~{\small($\pm$0.041)} & 0.089~{\small($\pm$0.047)} & 0.097~{\small($\pm$0.055)} & 0.129~{\small($\pm$0.075)} & 0.068~{\small($\pm$0.027)} \\

    \quad GenMC & 0.069~{\small($\pm$0.019)} & 0.107~{\small($\pm$0.046)} & 0.113~{\small($\pm$0.060)} & 0.121~{\small($\pm$0.053)} & 0.072~{\small($\pm$0.027)} \\
    \hline
    T5$_\textsc{Large}$ \\
    \quad Text2Text$_\text{vanilla}$ & 0.077~{\small($\pm$0.008)} &  0.083~{\small($\pm$0.012)} & 0.081~{\small($\pm$0.012)} & 0.084~{\small($\pm$0.014)} & 0.078~{\small($\pm$0.011)} \\
    
    \quad UnifiedQA &
    0.108~{\small($\pm$0.037)} & 0.178~{\small($\pm$0.096)} & 0.190~{\small($\pm$0.107)} & 0.257~{\small($\pm$0.127)} & 0.130~{\small($\pm$0.052)} \\
    
    \quad GenMC & 0.105~{\small($\pm$0.027)} & 0.178~{\small($\pm$0.078)} & 0.219~{\small($\pm$0.120)} & 0.242~{\small($\pm$0.112)} & 0.127~{\small($\pm$0.048)} \\
    \hline
    
\end{tabular}
\caption{Inference time for answering a question~(seconds).}
\label{table:run_time}
\end{table*}

To ensure the annotation quality, we aggregate annotated results from three students for every dataset using majority vote. If all three students annotate differently from each other for an instance, we introduce a fourth student to arbitrate.

\begin{figure}[t!]
    \centering
    \includegraphics[width=0.5\textwidth]{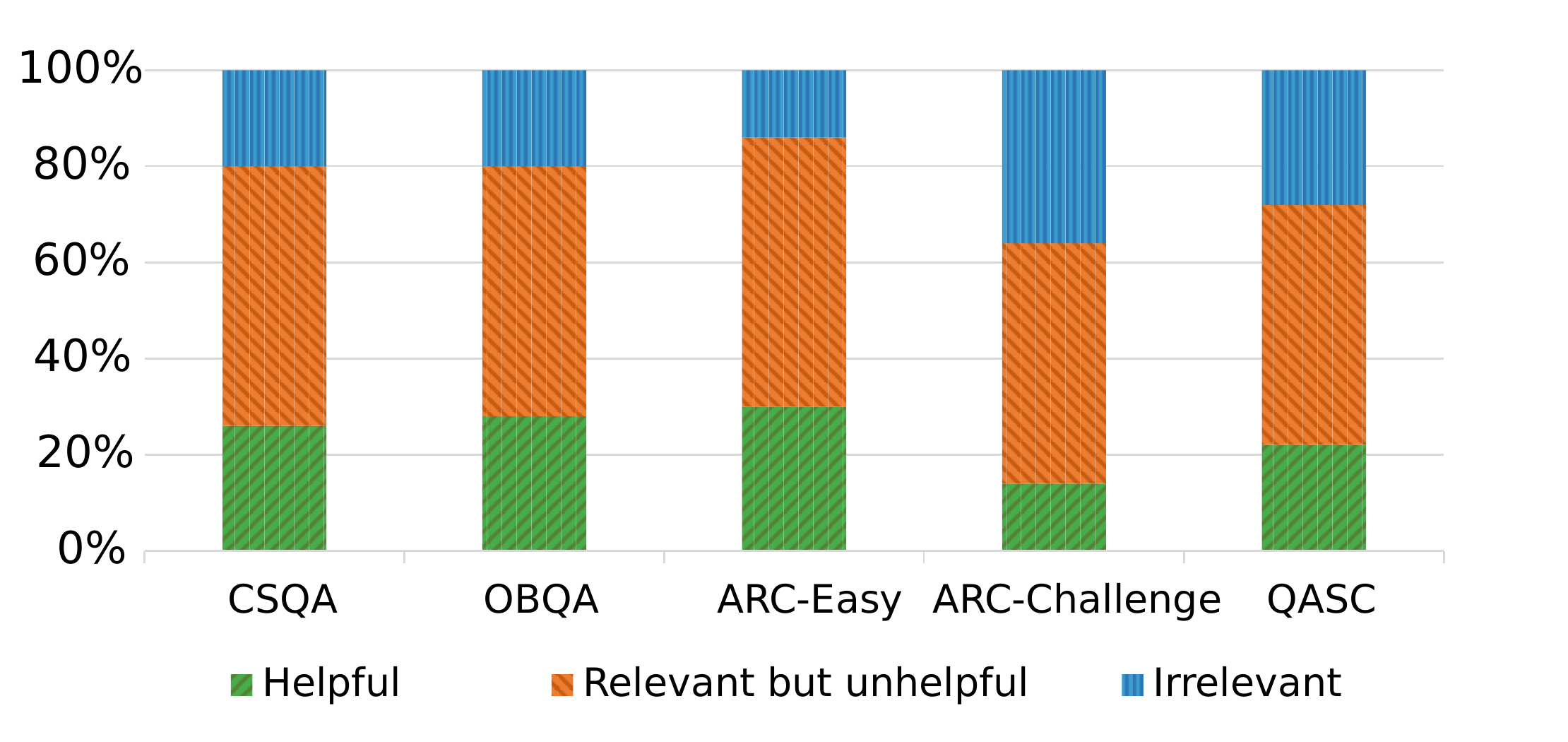}
    \caption{Distribution of clue types in negative cases on each dataset.}
    \label{fig:negFig}
\end{figure}

Table~\ref{table:negExp} shows the percent of each clue type across all datasets with an example for each type. Figure~\ref{fig:negFig} breaks down by dataset. Though the majority of our clues are relevant (i.e., 76.4\% of them are relevant across all datasets), which seems positive, only 24\% of the clues are deemed as helpful. This suggests a great room for improvement. In our future research, we will focus on how to generate more helpful clues from questions.


\subsection{Inference Time and Model Size}

Table~\ref{table:run_time} shows the inference time for answering a question.
GenMC is slower than Text2Text$_\text{vanilla}$, but their inference time has the same scale, suggesting that GenMC is more cost-effective considering its superior accuracy. GenMC and UnifiedQA are comparable in inference time.


Among T5$_\textsc{Base}$ based models, Text2Text$_\text{vanilla}$ and UnifiedQA have 223~M~parameters, while GenMC is slightly larger with 234~M parameters. Among T5$_\textsc{Large}$ based models, Text2Text$_\text{vanilla}$ and UnifiedQA have 738~M~parameters, while GenMC has 757~M~parameters.

\section{Conclusion}
\label{sec:conclusion}
We present GenMC, a simple yet effective model which tailors pre-trained encoder-decoders for MCQA tasks. Compared with existing usages of pre-trained encoder-decoders for MCQA, our model fully exploits the pre-trained encoder-decoders' NLG capabilities to generate a clue from the input question, which facilitates deep understanding of question-option pairs. Experimental results further verify the superiority of GenMC over existing usages. Notably, our model achieves promising results without using any provided documents or external corpora, showing an interesting application of PLMs by directly inducing either commonsense or scientific knowledge from them through clue generation.

In the future, we will focus on how to further improve the clue generation quality, which remains a bottleneck of GenMC. We hope this work will spur more research in how to better use pre-trained encoder-decoders for not only MCQA, but also beyond; for tasks with divergent structures from the pre-training, a smarter use of PLMs can boost the performance significantly.  


\section*{Acknowledgments}
This work was supported in part by the NSFC (62072224) and in part by the Beijing Academy of Artificial Intelligence (BAAI).

\bibliography{custom}

\begin{thebibliography}{38}
\expandafter\ifx\csname natexlab\endcsname\relax\def\natexlab#1{#1}\fi

\bibitem[{Chen et~al.(2019)Chen, Cui, Ma, Wang, and Hu}]{CSA}
Zhipeng Chen, Yiming Cui, Wentao Ma, Shijin Wang, and Guoping Hu. 2019.
\newblock \href {https://doi.org/10.1609/aaai.v33i01.33016276} {Convolutional
  spatial attention model for reading comprehension with multiple-choice
  questions}.
\newblock In \emph{The Thirty-Third {AAAI} Conference on Artificial
  Intelligence, {AAAI} 2019, The Thirty-First Innovative Applications of
  Artificial Intelligence Conference, {IAAI} 2019, The Ninth {AAAI} Symposium
  on Educational Advances in Artificial Intelligence, {EAAI} 2019, Honolulu,
  Hawaii, USA, January 27 - February 1, 2019}, pages 6276--6283. {AAAI} Press.

\bibitem[{Clark et~al.(2018)Clark, Cowhey, Etzioni, Khot, Sabharwal, Schoenick,
  and Tafjord}]{ARC}
Peter Clark, Isaac Cowhey, Oren Etzioni, Tushar Khot, Ashish Sabharwal, Carissa
  Schoenick, and Oyvind Tafjord. 2018.
\newblock \href {http://arxiv.org/abs/1803.05457} {Think you have solved
  question answering? try arc, the {AI2} reasoning challenge}.
\newblock \emph{CoRR}, abs/1803.05457.

\bibitem[{Devlin et~al.(2019)Devlin, Chang, Lee, and Toutanova}]{BERT}
Jacob Devlin, Ming{-}Wei Chang, Kenton Lee, and Kristina Toutanova. 2019.
\newblock \href {https://doi.org/10.18653/v1/n19-1423} {{BERT:} pre-training of
  deep bidirectional transformers for language understanding}.
\newblock In \emph{Proceedings of the 2019 Conference of the North American
  Chapter of the Association for Computational Linguistics: Human Language
  Technologies, {NAACL-HLT} 2019, Minneapolis, MN, USA, June 2-7, 2019, Volume
  1 (Long and Short Papers)}, pages 4171--4186. Association for Computational
  Linguistics.

\bibitem[{Huang et~al.(2019)Huang, Shen, Li, Wei, Cheng, Zhou, Dai, and
  Qu}]{DBLP:conf/emnlp/HuangSLWCZDQ19}
Zixian Huang, Yulin Shen, Xiao Li, Yuang Wei, Gong Cheng, Lin Zhou, Xinyu Dai,
  and Yuzhong Qu. 2019.
\newblock \href {https://doi.org/10.18653/v1/D19-1597} {Geosqa: {A} benchmark
  for scenario-based question answering in the geography domain at high school
  level}.
\newblock In \emph{Proceedings of the 2019 Conference on Empirical Methods in
  Natural Language Processing and the 9th International Joint Conference on
  Natural Language Processing, {EMNLP-IJCNLP} 2019, Hong Kong, China, November
  3-7, 2019}, pages 5865--5870. Association for Computational Linguistics.

\bibitem[{Huang et~al.(2021)Huang, Wu, Shen, Cheng, and Qu}]{JEEVES}
Zixian Huang, Ao~Wu, Yulin Shen, Gong Cheng, and Yuzhong Qu. 2021.
\newblock \href {https://aclanthology.org/2021.findings-emnlp.84} {When
  retriever-reader meets scenario-based multiple-choice questions}.
\newblock In \emph{Findings of the Association for Computational Linguistics:
  {EMNLP} 2021, Virtual Event / Punta Cana, Dominican Republic, 16-20 November,
  2021}, pages 985--994. Association for Computational Linguistics.

\bibitem[{Jiang et~al.(2020)Jiang, Xu, Araki, and Neubig}]{What_LM_Know}
Zhengbao Jiang, Frank~F. Xu, Jun Araki, and Graham Neubig. 2020.
\newblock \href {https://transacl.org/ojs/index.php/tacl/article/view/1983}
  {How can we know what language models know}.
\newblock \emph{Trans. Assoc. Comput. Linguistics}, 8:423--438.

\bibitem[{Khashabi et~al.(2020)Khashabi, Min, Khot, Sabharwal, Tafjord, Clark,
  and Hajishirzi}]{UnifiedQA}
Daniel Khashabi, Sewon Min, Tushar Khot, Ashish Sabharwal, Oyvind Tafjord,
  Peter Clark, and Hannaneh Hajishirzi. 2020.
\newblock \href {https://doi.org/10.18653/v1/2020.findings-emnlp.171}
  {Unifiedqa: Crossing format boundaries with a single {QA} system}.
\newblock In \emph{Findings of the Association for Computational Linguistics:
  {EMNLP} 2020, Online Event, 16-20 November 2020}, volume {EMNLP} 2020 of
  \emph{Findings of {ACL}}, pages 1896--1907. Association for Computational
  Linguistics.

\bibitem[{Khot et~al.(2020)Khot, Clark, Guerquin, Jansen, and Sabharwal}]{QASC}
Tushar Khot, Peter Clark, Michal Guerquin, Peter Jansen, and Ashish Sabharwal.
  2020.
\newblock \href {https://aaai.org/ojs/index.php/AAAI/article/view/6319}
  {{QASC:} {A} dataset for question answering via sentence composition}.
\newblock In \emph{The Thirty-Fourth {AAAI} Conference on Artificial
  Intelligence, {AAAI} 2020, The Thirty-Second Innovative Applications of
  Artificial Intelligence Conference, {IAAI} 2020, The Tenth {AAAI} Symposium
  on Educational Advances in Artificial Intelligence, {EAAI} 2020, New York,
  NY, USA, February 7-12, 2020}, pages 8082--8090. {AAAI} Press.

\bibitem[{Khot et~al.(2019)Khot, Sabharwal, and Clark}]{GapQA}
Tushar Khot, Ashish Sabharwal, and Peter Clark. 2019.
\newblock \href {https://doi.org/10.18653/v1/D19-1281} {What's missing: {A}
  knowledge gap guided approach for multi-hop question answering}.
\newblock In \emph{Proceedings of the 2019 Conference on Empirical Methods in
  Natural Language Processing and the 9th International Joint Conference on
  Natural Language Processing, {EMNLP-IJCNLP} 2019, Hong Kong, China, November
  3-7, 2019}, pages 2814--2828. Association for Computational Linguistics.

\bibitem[{Lan et~al.(2020)Lan, Chen, Goodman, Gimpel, Sharma, and
  Soricut}]{ALBert}
Zhenzhong Lan, Mingda Chen, Sebastian Goodman, Kevin Gimpel, Piyush Sharma, and
  Radu Soricut. 2020.
\newblock \href {https://openreview.net/forum?id=H1eA7AEtvS} {{ALBERT:} {A}
  lite {BERT} for self-supervised learning of language representations}.
\newblock In \emph{8th International Conference on Learning Representations,
  {ICLR} 2020, Addis Ababa, Ethiopia, April 26-30, 2020}. OpenReview.net.

\bibitem[{Latcinnik and Berant(2020)}]{ExplainQA}
Veronica Latcinnik and Jonathan Berant. 2020.
\newblock \href {http://arxiv.org/abs/2004.05569} {Explaining question
  answering models through text generation}.
\newblock \emph{CoRR}, abs/2004.05569.

\bibitem[{Lewis et~al.(2020)Lewis, Liu, Goyal, Ghazvininejad, Mohamed, Levy,
  Stoyanov, and Zettlemoyer}]{BART}
Mike Lewis, Yinhan Liu, Naman Goyal, Marjan Ghazvininejad, Abdelrahman Mohamed,
  Omer Levy, Veselin Stoyanov, and Luke Zettlemoyer. 2020.
\newblock \href {https://doi.org/10.18653/v1/2020.acl-main.703} {{BART:}
  denoising sequence-to-sequence pre-training for natural language generation,
  translation, and comprehension}.
\newblock In \emph{Proceedings of the 58th Annual Meeting of the Association
  for Computational Linguistics, {ACL} 2020, Online, July 5-10, 2020}, pages
  7871--7880. Association for Computational Linguistics.

\bibitem[{Li et~al.(2022)Li, Cheng, Chen, Sun, and
  Qu}]{DBLP:journals/corr/abs-2203-08992}
Xiao Li, Gong Cheng, Ziheng Chen, Yawei Sun, and Yuzhong Qu. 2022.
\newblock \href {https://doi.org/10.48550/arXiv.2203.08992} {Adalogn: Adaptive
  logic graph network for reasoning-based machine reading comprehension}.
\newblock \emph{CoRR}, abs/2203.08992.

\bibitem[{Li et~al.(2021)Li, Sun, and Cheng}]{DBLP:conf/aaai/LiS021}
Xiao Li, Yawei Sun, and Gong Cheng. 2021.
\newblock \href {https://ojs.aaai.org/index.php/AAAI/article/view/17570}
  {{TSQA:} tabular scenario based question answering}.
\newblock In \emph{Thirty-Fifth {AAAI} Conference on Artificial Intelligence,
  {AAAI} 2021, Thirty-Third Conference on Innovative Applications of Artificial
  Intelligence, {IAAI} 2021, The Eleventh Symposium on Educational Advances in
  Artificial Intelligence, {EAAI} 2021, Virtual Event, February 2-9, 2021},
  pages 13297--13305. {AAAI} Press.

\bibitem[{Liu et~al.(2020{\natexlab{a}})Liu, Gong, Fu, Yan, Chen, Jiang, Lv,
  and Duan}]{RikiNet}
Dayiheng Liu, Yeyun Gong, Jie Fu, Yu~Yan, Jiusheng Chen, Daxin Jiang, Jiancheng
  Lv, and Nan Duan. 2020{\natexlab{a}}.
\newblock {RikiNet:} reading {Wikipedia} pages for natural question answering.
\newblock In \emph{{ACL}}, pages 6762--6771.

\bibitem[{Liu et~al.(2021)Liu, Shakeri, Yu, and Li}]{EncT5}
Frederick Liu, Siamak Shakeri, Hongkun Yu, and Jing Li. 2021.
\newblock \href {http://arxiv.org/abs/2110.08426} {Enct5: Fine-tuning {T5}
  encoder for non-autoregressive tasks}.
\newblock \emph{CoRR}, abs/2110.08426.

\bibitem[{Liu et~al.(2020{\natexlab{b}})Liu, Cui, Liu, Huang, Wang, and
  Zhang}]{DBLP:conf/ijcai/LiuCLHWZ20}
Jian Liu, Leyang Cui, Hanmeng Liu, Dandan Huang, Yile Wang, and Yue Zhang.
  2020{\natexlab{b}}.
\newblock \href {https://doi.org/10.24963/ijcai.2020/501} {Logiqa: {A}
  challenge dataset for machine reading comprehension with logical reasoning}.
\newblock In \emph{Proceedings of the Twenty-Ninth International Joint
  Conference on Artificial Intelligence, {IJCAI} 2020}, pages 3622--3628.
  ijcai.org.

\bibitem[{Liu et~al.(2020{\natexlab{c}})Liu, Yang, You, Fan, and Yu}]{CEGI}
Ye~Liu, Tao Yang, Zeyu You, Wei Fan, and Philip~S. Yu. 2020{\natexlab{c}}.
\newblock \href {https://aclanthology.org/2020.sigdial-1.9/} {Commonsense
  evidence generation and injection in reading comprehension}.
\newblock In \emph{Proceedings of the 21th Annual Meeting of the Special
  Interest Group on Discourse and Dialogue, SIGdial 2020, 1st virtual meeting,
  July 1-3, 2020}, pages 61--73. Association for Computational Linguistics.

\bibitem[{Liu et~al.(2019)Liu, Ott, Goyal, Du, Joshi, Chen, Levy, Lewis,
  Zettlemoyer, and Stoyanov}]{RoBERTa}
Yinhan Liu, Myle Ott, Naman Goyal, Jingfei Du, Mandar Joshi, Danqi Chen, Omer
  Levy, Mike Lewis, Luke Zettlemoyer, and Veselin Stoyanov. 2019.
\newblock \href {http://arxiv.org/abs/1907.11692} {Roberta: {A} robustly
  optimized {BERT} pretraining approach}.
\newblock \emph{CoRR}, abs/1907.11692.

\bibitem[{Mao et~al.(2021)Mao, He, Liu, Shen, Gao, Han, and Chen}]{GAR}
Yuning Mao, Pengcheng He, Xiaodong Liu, Yelong Shen, Jianfeng Gao, Jiawei Han,
  and Weizhu Chen. 2021.
\newblock \href {https://doi.org/10.18653/v1/2021.acl-long.316}
  {Generation-augmented retrieval for open-domain question answering}.
\newblock In \emph{Proceedings of the 59th Annual Meeting of the Association
  for Computational Linguistics and the 11th International Joint Conference on
  Natural Language Processing, {ACL/IJCNLP} 2021, (Volume 1: Long Papers),
  Virtual Event, August 1-6, 2021}, pages 4089--4100. Association for
  Computational Linguistics.

\bibitem[{Mihaylov et~al.(2018)Mihaylov, Clark, Khot, and Sabharwal}]{OBQA}
Todor Mihaylov, Peter Clark, Tushar Khot, and Ashish Sabharwal. 2018.
\newblock \href {https://doi.org/10.18653/v1/d18-1260} {Can a suit of armor
  conduct electricity? {A} new dataset for open book question answering}.
\newblock In \emph{Proceedings of the 2018 Conference on Empirical Methods in
  Natural Language Processing, Brussels, Belgium, October 31 - November 4,
  2018}, pages 2381--2391. Association for Computational Linguistics.

\bibitem[{Petroni et~al.(2019)Petroni, Rockt{\"{a}}schel, Riedel, Lewis,
  Bakhtin, Wu, and Miller}]{LM_AS_KB}
Fabio Petroni, Tim Rockt{\"{a}}schel, Sebastian Riedel, Patrick S.~H. Lewis,
  Anton Bakhtin, Yuxiang Wu, and Alexander~H. Miller. 2019.
\newblock \href {https://doi.org/10.18653/v1/D19-1250} {Language models as
  knowledge bases?}
\newblock In \emph{Proceedings of the 2019 Conference on Empirical Methods in
  Natural Language Processing and the 9th International Joint Conference on
  Natural Language Processing, {EMNLP-IJCNLP} 2019, Hong Kong, China, November
  3-7, 2019}, pages 2463--2473. Association for Computational Linguistics.

\bibitem[{Raffel et~al.(2020)Raffel, Shazeer, Roberts, Lee, Narang, Matena,
  Zhou, Li, and Liu}]{T5}
Colin Raffel, Noam Shazeer, Adam Roberts, Katherine Lee, Sharan Narang, Michael
  Matena, Yanqi Zhou, Wei Li, and Peter~J. Liu. 2020.
\newblock \href {http://jmlr.org/papers/v21/20-074.html} {Exploring the limits
  of transfer learning with a unified text-to-text transformer}.
\newblock \emph{J. Mach. Learn. Res.}, 21:140:1--140:67.

\bibitem[{Rajani et~al.(2019)Rajani, McCann, Xiong, and Socher}]{CoS-E}
Nazneen~Fatema Rajani, Bryan McCann, Caiming Xiong, and Richard Socher. 2019.
\newblock \href {https://doi.org/10.18653/v1/p19-1487} {Explain yourself!
  leveraging language models for commonsense reasoning}.
\newblock In \emph{Proceedings of the 57th Conference of the Association for
  Computational Linguistics, {ACL} 2019, Florence, Italy, July 28- August 2,
  2019, Volume 1: Long Papers}, pages 4932--4942. Association for Computational
  Linguistics.

\bibitem[{Ran et~al.(2019)Ran, Li, Hu, and Zhou}]{OCN}
Qiu Ran, Peng Li, Weiwei Hu, and Jie Zhou. 2019.
\newblock \href {http://arxiv.org/abs/1903.03033} {Option comparison network
  for multiple-choice reading comprehension}.
\newblock \emph{CoRR}, abs/1903.03033.

\bibitem[{Shao et~al.(2021)Shao, Geng, Liu, Dai, Yang, Zhe, Bao, and Qiu}]{CPT}
Yunfan Shao, Zhichao Geng, Yitao Liu, Junqi Dai, Fei Yang, Li~Zhe, Hujun Bao,
  and Xipeng Qiu. 2021.
\newblock \href {http://arxiv.org/abs/2109.05729} {{CPT:} {A} pre-trained
  unbalanced transformer for both chinese language understanding and
  generation}.
\newblock \emph{CoRR}, abs/2109.05729.

\bibitem[{Shin et~al.(2020)Shin, Razeghi, IV, Wallace, and Singh}]{AutoPrompt}
Taylor Shin, Yasaman Razeghi, Robert L.~Logan IV, Eric Wallace, and Sameer
  Singh. 2020.
\newblock \href {https://doi.org/10.18653/v1/2020.emnlp-main.346} {Autoprompt:
  Eliciting knowledge from language models with automatically generated
  prompts}.
\newblock In \emph{Proceedings of the 2020 Conference on Empirical Methods in
  Natural Language Processing, {EMNLP} 2020, Online, November 16-20, 2020},
  pages 4222--4235. Association for Computational Linguistics.

\bibitem[{Shwartz et~al.(2020)Shwartz, West, Bras, Bhagavatula, and
  Choi}]{QG-Self-Talk}
Vered Shwartz, Peter West, Ronan~Le Bras, Chandra Bhagavatula, and Yejin Choi.
  2020.
\newblock \href {https://doi.org/10.18653/v1/2020.emnlp-main.373} {Unsupervised
  commonsense question answering with self-talk}.
\newblock In \emph{Proceedings of the 2020 Conference on Empirical Methods in
  Natural Language Processing, {EMNLP} 2020, Online, November 16-20, 2020},
  pages 4615--4629. Association for Computational Linguistics.

\bibitem[{Sun et~al.(2019)Sun, Yu, Yu, and Cardie}]{MRC_Strategy}
Kai Sun, Dian Yu, Dong Yu, and Claire Cardie. 2019.
\newblock \href {https://doi.org/10.18653/v1/n19-1270} {Improving machine
  reading comprehension with general reading strategies}.
\newblock In \emph{Proceedings of the 2019 Conference of the North American
  Chapter of the Association for Computational Linguistics: Human Language
  Technologies, {NAACL-HLT} 2019, Minneapolis, MN, USA, June 2-7, 2019, Volume
  1 (Long and Short Papers)}, pages 2633--2643. Association for Computational
  Linguistics.

\bibitem[{Sun et~al.(2021)Sun, Wang, Feng, Ding, Pang, Shang, Liu, Chen, Zhao,
  Lu, Liu, Wu, Gong, Liang, Shang, Sun, Liu, Ouyang, Yu, Tian, Wu, and
  Wang}]{ERNIE3}
Yu~Sun, Shuohuan Wang, Shikun Feng, Siyu Ding, Chao Pang, Junyuan Shang,
  Jiaxiang Liu, Xuyi Chen, Yanbin Zhao, Yuxiang Lu, Weixin Liu, Zhihua Wu,
  Weibao Gong, Jianzhong Liang, Zhizhou Shang, Peng Sun, Wei Liu, Xuan Ouyang,
  Dianhai Yu, Hao Tian, Hua Wu, and Haifeng Wang. 2021.
\newblock \href {http://arxiv.org/abs/2107.02137} {{ERNIE} 3.0: Large-scale
  knowledge enhanced pre-training for language understanding and generation}.
\newblock \emph{CoRR}, abs/2107.02137.

\bibitem[{Talmor et~al.(2019)Talmor, Herzig, Lourie, and Berant}]{CSQA}
Alon Talmor, Jonathan Herzig, Nicholas Lourie, and Jonathan Berant. 2019.
\newblock \href {https://doi.org/10.18653/v1/n19-1421} {Commonsenseqa: {A}
  question answering challenge targeting commonsense knowledge}.
\newblock In \emph{Proceedings of the 2019 Conference of the North American
  Chapter of the Association for Computational Linguistics: Human Language
  Technologies, {NAACL-HLT} 2019, Minneapolis, MN, USA, June 2-7, 2019, Volume
  1 (Long and Short Papers)}, pages 4149--4158. Association for Computational
  Linguistics.

\bibitem[{Tang et~al.(2019)Tang, Cai, and Zhuo}]{MMN}
Min Tang, Jiaran Cai, and Hankz~Hankui Zhuo. 2019.
\newblock \href {https://doi.org/10.1609/aaai.v33i01.33017088} {Multi-matching
  network for multiple choice reading comprehension}.
\newblock In \emph{The Thirty-Third {AAAI} Conference on Artificial
  Intelligence, {AAAI} 2019, The Thirty-First Innovative Applications of
  Artificial Intelligence Conference, {IAAI} 2019, The Ninth {AAAI} Symposium
  on Educational Advances in Artificial Intelligence, {EAAI} 2019, Honolulu,
  Hawaii, USA, January 27 - February 1, 2019}, pages 7088--7095. {AAAI} Press.

\bibitem[{Vaswani et~al.(2017)Vaswani, Shazeer, Parmar, Uszkoreit, Jones,
  Gomez, Kaiser, and Polosukhin}]{transformer}
Ashish Vaswani, Noam Shazeer, Niki Parmar, Jakob Uszkoreit, Llion Jones,
  Aidan~N. Gomez, Lukasz Kaiser, and Illia Polosukhin. 2017.
\newblock \href
  {https://proceedings.neurips.cc/paper/2017/hash/3f5ee243547dee91fbd053c1c4a845aa-Abstract.html}
  {Attention is all you need}.
\newblock In \emph{Advances in Neural Information Processing Systems 30: Annual
  Conference on Neural Information Processing Systems 2017, December 4-9, 2017,
  Long Beach, CA, {USA}}, pages 5998--6008.

\bibitem[{Yu et~al.(2020)Yu, Jiang, Dong, and Feng}]{DBLP:conf/iclr/YuJDF20}
Weihao Yu, Zihang Jiang, Yanfei Dong, and Jiashi Feng. 2020.
\newblock \href {https://openreview.net/forum?id=HJgJtT4tvB} {Reclor: {A}
  reading comprehension dataset requiring logical reasoning}.
\newblock In \emph{8th International Conference on Learning Representations,
  {ICLR} 2020, Addis Ababa, Ethiopia, April 26-30, 2020}. OpenReview.net.

\bibitem[{Zellers et~al.(2019)Zellers, Holtzman, Bisk, Farhadi, and
  Choi}]{zellers-etal-2019-hellaswag}
Rowan Zellers, Ari Holtzman, Yonatan Bisk, Ali Farhadi, and Yejin Choi. 2019.
\newblock \href {https://doi.org/10.18653/v1/P19-1472} {{H}ella{S}wag: Can a
  machine really finish your sentence?}
\newblock In \emph{Proceedings of the 57th Annual Meeting of the Association
  for Computational Linguistics}, pages 4791--4800, Florence, Italy.
  Association for Computational Linguistics.

\bibitem[{Zhang et~al.(2020)Zhang, Zhao, Wu, Zhang, Zhou, and Zhou}]{DCMN}
Shuailiang Zhang, Hai Zhao, Yuwei Wu, Zhuosheng Zhang, Xi~Zhou, and Xiang Zhou.
  2020.
\newblock \href {https://aaai.org/ojs/index.php/AAAI/article/view/6502}
  {{DCMN+:} dual co-matching network for multi-choice reading comprehension}.
\newblock In \emph{The Thirty-Fourth {AAAI} Conference on Artificial
  Intelligence, {AAAI} 2020, The Thirty-Second Innovative Applications of
  Artificial Intelligence Conference, {IAAI} 2020, The Tenth {AAAI} Symposium
  on Educational Advances in Artificial Intelligence, {EAAI} 2020, New York,
  NY, USA, February 7-12, 2020}, pages 9563--9570. {AAAI} Press.

\bibitem[{Zhou et~al.(2021)Zhou, Lee, Selvam, Lee, and Ren}]{CALM}
Wangchunshu Zhou, Dong{-}Ho Lee, Ravi~Kiran Selvam, Seyeon Lee, and Xiang Ren.
  2021.
\newblock \href {https://openreview.net/forum?id=3k20LAiHYL2} {Pre-training
  text-to-text transformers for concept-centric common sense}.
\newblock In \emph{9th International Conference on Learning Representations,
  {ICLR} 2021, Virtual Event, Austria, May 3-7, 2021}. OpenReview.net.

\bibitem[{Zhu et~al.(2020)Zhu, Zhao, and Li}]{DUMA}
Pengfei Zhu, Hai Zhao, and Xiaoguang Li. 2020.
\newblock \href {http://arxiv.org/abs/2001.09415} {Dual multi-head co-attention
  for multi-choice reading comprehension}.
\newblock \emph{CoRR}, abs/2001.09415.

\end{thebibliography}
\bibliographystyle{acl_natbib}

\end{document}